\documentclass[letterpaper]{article} 
\usepackage{aaai2026}  
\usepackage{times}  
\usepackage{helvet}  
\usepackage{courier}  
\usepackage[hyphens]{url}  
\usepackage{graphicx} 
\usepackage{natbib}  
\usepackage{caption} 
\usepackage{algorithm}
\usepackage{algorithmic}

\usepackage{amssymb}
\usepackage{amsmath}

\usepackage{xcolor}
\usepackage{afterpage} 
\usepackage{array}
\usepackage{booktabs}
\usepackage{newfloat}
\usepackage{listings}
\DeclareCaptionStyle{ruled}{labelfont=normalfont,labelsep=colon,strut=off} 
\floatstyle{ruled}
\newfloat{listing}{tb}{lst}{}
\floatname{listing}{Listing}
\title{Target-Balanced Score Distillation}

\author{
    Zhou Xu\textsuperscript{\rm 1}\equalcontrib, 
    Qi Wang\textsuperscript{\rm 2}\equalcontrib, 
    Yuxiao Yang\textsuperscript{\rm 1}, 
    Luyuan Zhang\textsuperscript{\rm 1}, 
    Zhang Liang\textsuperscript{\rm 2 \textdagger}, 
    Yang Li\textsuperscript{\rm 1}\thanks{Corresponding author.}
}
\affiliations{
    \textsuperscript{\rm 1}Tsinghua University\\
    \textsuperscript{\rm 2}Xidian University\\
    xu-z25@mails.tsinghua.edu.cn
}

\usepackage{bibentry}

\begin{document}
\maketitle

\begin{figure*}[ht!]  
\centering
\includegraphics[width=1\textwidth]{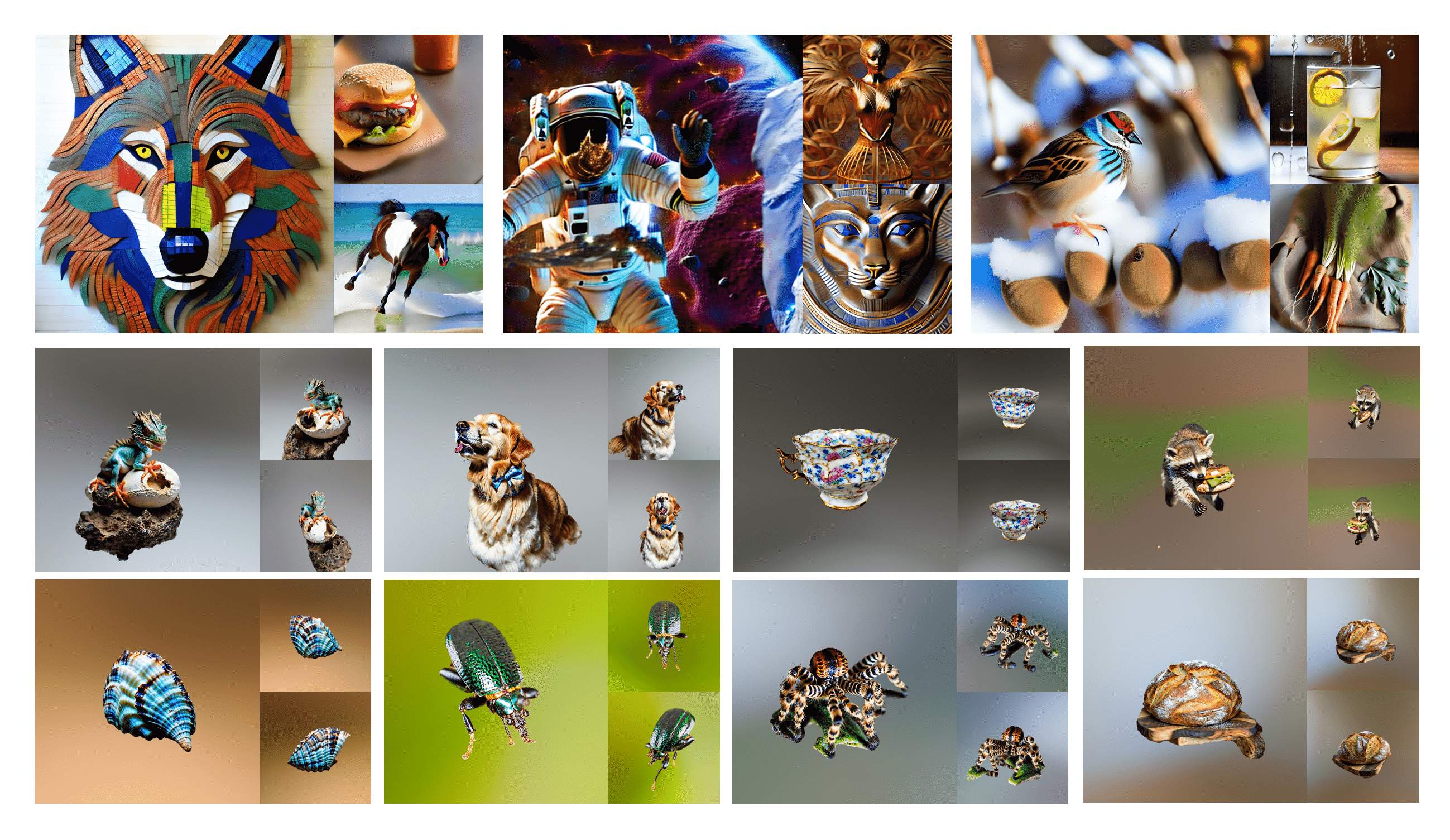} %
\caption{Results obtained with our Target-Balanced Score Distillation (TBSD). Top: a gallery of images optimized with TBSD. Bottom: two rows NeRFs generated by TBSD (other examples are included in the supplementary material).} %
\label{fig:teaser} %
\end{figure*}

\begin{abstract}
Score Distillation Sampling (SDS) enables 3D asset generation by distilling priors from pretrained 2D text-to-image diffusion models, but vanilla SDS suffers from over-saturation and over-smoothing. To mitigate this issue, recent variants have incorporated negative prompts. However, these methods face a critical trade-off: limited texture optimization, or significant texture gains with shape distortion. In this work, we first conduct a systematic analysis and reveal that this trade-off is fundamentally governed by the utilization of the negative prompts, where \textbf{Target Negative Prompts (TNP)} that embed target information in the negative prompts dramatically enhancing texture realism and fidelity but inducing shape distortions.
Informed by this key insight, we introduce the \textbf{Target-Balanced Score Distillation (TBSD)}. It formulates generation as a multi-objective optimization problem and introduces an adaptive strategy that effectively resolves the aforementioned trade-off. Extensive experiments demonstrate that TBSD significantly outperforms existing state-of-the-art methods, yielding 3D assets with high-fidelity textures and geometrically accurate shape. 
\end{abstract}

\begin{links}
    \link{Code}{https://github.com/XiaocatMomo/TBSD}
\end{links}

\section{Introduction}

In recent years, text-to-image diffusion models have made remarkable progress, significantly advancing the field of image synthesis \cite{t2i}. These models \cite{dataset}, typically trained on large-scale datasets and powered by massive parameter capacities, are capable of generating highly realistic and detail-rich images from simple textual descriptions. However, achieving comparable generation quality remains technically challenging in scenarios with relatively scarce training data, such as 3D generation. The high demand for 3D objects in downstream tasks like computer graphics and robotics has led to the development of methods such as Score Distillation Sampling (SDS) \cite{DreamFusion}, which enables knowledge transfer from pre-trained 2D diffusion models to 3D content. The core principle of SDS is to optimize 3D representations \cite{3DGS,nerf,instant} such that their rendered images approach high-probability density regions under text conditions, with supervision provided by pre-trained 2D diffusion models. Due to this formulation, these SDS-based methods \cite{DreamGaussian,Fantasia3D,GaussianDreamer,HIFA,ProlificDreamer,SDI,LucidDreamer,PlacidDreamer} do not require 3D data for training and can generate 3D results from various text prompts. 

Despite its success, existing studies \cite{LucidDreamer} have noted that SDS exhibits an averaging effect during generation, leading to color over-saturation and overly smooth textures. Inspired by the success of negative prompts in 2D diffusion models, negative prompts have achieved success in guiding models by specifying ``content not to generate" \cite{under_neg,neg1}. Thus, some studies have attempted to introduce negative prompts into score distillation to alleviate the above issues. Existing methods typically leverage negative prompts either as approximate domain correction terms during early training \cite{NFSD, CSD, Bridge}, as an auxiliary acceleration mechanism \cite{CSD}, or to guide texture optimization via source distribution estimation \cite{NFSD, Bridge}. However, these methods suffer from a critical trade-off: either texture optimization remains limited, or significant improvements in texture fidelity come at the cost of shape distortion.

In this paper, we first conduct a comprehensive review of existing SDS variants that incorporate negative prompts and reveal that this trade-off is fundamentally governed by the utilization of the negative prompts. \emph{Interestingly, Our analysis finds that \textbf{Target Negative Prompts (TNP)}, which embed only the target information in the negative prompts, can dramatically enhancing the realism and fidelity of generated textures.} However, this overly focused guidance often leads to loss of target information and subsequent global shape distortions, which also reported in prior work \cite{Bridge}.

To address this limitation, we further propose a novel score distillation framework termed \textbf{Target-Balanced Score Distillation (TBSD)}. Our method formulates SDS as a multi-objective optimization problem: one objective (shape guidance) is optimized using the classifier-free guidance term of standard SDS branch, while the other objective (texture enhancement) is guided by the TNP branch. It introduces an adaptive strategy ensuring that TBSD initially focuses more on shape optimization, enabling the model to generate an accurate shape in the early stages. As training progresses, the optimization gradually shifts toward the texture objective, with TBSD maintaining a good balance between shape and texture to allow texture quality to improve while preserving essential target information. This dynamic balancing strategy effectively resolves the aforementioned trade-off. As shown in Figure \ref{fig:teaser}, our TBSD method can produce 3D assets that preserve geometric correctness while exhibiting extremely realistic and detailed textures. 

Our key contributions are as follows:
\begin{itemize}

    \item We conduct a detailed analysis of existing SDS variants that incorporate negative prompts. We identify and explain the mechanism by which Target Negative Prompts (TNP) enhance texture fidelity while potentially shape distortion.

    \item To mitigate the shape distortion introduced by TNP, we propose Target-Balanced Score Distillation (TBSD), a dynamic optimization framework that simultaneously balances shape preservation and texture realism.

    \item Extensive experiments across both 2D and 3D generation demonstrate that TBSD outperforms state-of-the-art methods, achieving the generation with high-fidelity textures and geometrically accurate shape.

\end{itemize}

\section{Related Works}
\textbf{Text-to-3D Generation.} DreamFields \cite{dreamfield} guides NeRF \cite{nerf} optimization using the pre-trained CLIP \cite{CLIP} model. DreamFusion \cite{DreamFusion} proposes Score Distillation Sampling (SDS), which enables generation by leveraging 2D diffusion models. Subsequent studies \cite{sweetdreamer,GSGEN,Fantasia3D,Magic3D,GaussianDreamer,dreamtime,richdreamer,HumanGaussian,HumanNorm} have improved SDS-based generation from multiple aspects. Given the high reliance on SDS, addressing its issues is crucial. SDS suffers from problems such as over-smoothing and over-saturation. Additionally, it requires a large conditional guidance scale, which further exacerbates over-saturation. Relevant improvements include: HiFA \cite{HIFA} enhances performance through an iterative process and the introduction of additional loss terms; ProlificDreamer \cite{ProlificDreamer} proposes Variational Score Distillation (VSD) to alleviate issues like over-saturation; Consistent3D \cite{Consistent3D} presents Consistency Distillation Sampling to reduce over-smoothing and other problems; LucidDreamer \cite{LucidDreamer} introduces Interval Score Matching (ISM) to optimize results and recent studies \cite{NFSD,CSD,Bridge,PlacidDreamer,lods,sds-lmc,cfd,isd} have also proposed other improvement methods for SDS. However, these methods have limitations, such as imposing significant computational burdens or yielding limited performance improvements.

\noindent\textbf{SDS with Negative Prompts.} Among SDS-related methods, NFSD \cite{NFSD} uses negative prompts to assist in extracting $\delta_{\mathrm{D}}$. For time steps $t \textgreater 200$, it approximates $\delta_{\mathrm{D}}$ by calculating the difference between noise under the null condition and noise under the negative text condition, and constructs a noise-free loss to improve generation quality without the need for a large guidance scale. CSD \cite{CSD} leverages negative prompts to achieve dual-objective optimization, driving the model to approach the target prompt and move away from negative states, thereby accelerating training and improving quality.
Bridge \cite{Bridge} uses negative prompts to represent the starting point of the Schrodinger bridge, alleviating the problem of inaccurate optimization starting points and enhancing the clarity of generated results. However, these methods fail to fully utilize the potential of negative text, with limited effectiveness or other issues such as shape distortion. This paper focuses on fully exploiting the advantages of negative prompts.

\begin{figure*}[ht!]  
\centering
\includegraphics[width=1\textwidth]{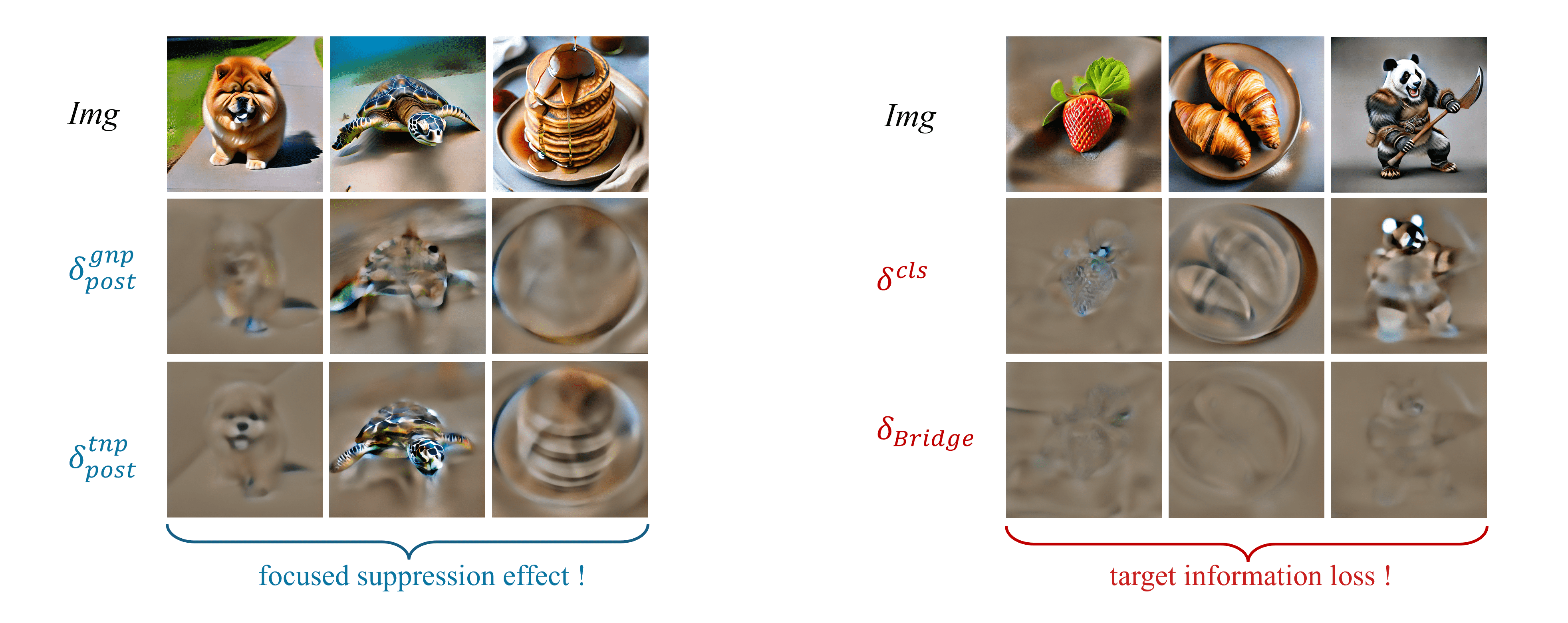} 
\caption{Visualization of $\delta_{\mathrm{post}}^{\mathrm{gnp}}, \ \delta_{\mathrm{post}}^{\mathrm{tnp}}, \ \delta^{\mathrm{cls}}$ and $\delta_{\mathrm{Bridge}}$. Top-row images are generated by TBSD. Visualization is done by decoding each $\delta$ with the VAE decoder of Stable Diffusion.} 
\label{fig:method2} 
\end{figure*}

\section{Analysis of Negative Prompts-based SDS Variants}
\subsection{Revisiting SDS and its variants}

\subsubsection{Score Distillation Sampling (SDS).}  
SDS leverages a pre-trained text-to-image diffusion model $\phi$ to guide the 3D representation parameterized by $\theta$. Specifically, for a given camera pose $\pi$, $x = g(\theta; \pi)$ represents the image rendered by the differentiable rendering function $g$. The SDS loss ensures that images obtained via $g$ from any viewpoint are aligned with the prompt $y$, and its form is as follows:
\begin{equation}
\nabla_\theta\mathcal{L}_{SDS}=\mathbf{E}_{t,\epsilon,\pi}[w(t)(\epsilon_\phi(\mathbf{x}_t;y,t)-\epsilon)\frac{\partial x}{\partial\theta}]
\label{eq:sds1}
\end{equation}
$w(t)$ serves as a weighting function, with $\mathbf{x}_t$ standing for a noisy variant of $\mathbf{x}$, which we denote as $\delta_{\mathrm{SDS}} := \epsilon_{\phi}\left(\mathbf{x}_{\theta,t};\varnothing,t\right)-\epsilon$. In practical implementation, Classifier-free guidance (CFG) is utilized in SDS. Specifically, $\delta_{\mathrm{SDS}}$ with CFG is expressed as:
\begin{equation}
\begin{aligned}
\delta_{\mathrm{SDS}}=& \epsilon_{\phi}\left(\mathbf{x}_{\theta,t};\varnothing,t\right)-\epsilon \ +\\
& s\cdot \underbrace{\left(\epsilon_{\phi}\left(\mathbf{x}_{\theta,t};y_{\mathrm{tgt}},t\right)-\epsilon_{\phi}\left(\mathbf{x}_{\theta,t};\varnothing,t\right)\right)}_{\delta^{\mathrm{cls}}}
\end{aligned}
\label{eq:sds2}
\end{equation}
\subsubsection{Noise Free Score Distillation (NFSD).}

\begin{equation}
\begin{aligned}
\delta_{\mathrm{NFSD}}=\  &s\cdot\left(\epsilon_{\phi}\left(\mathbf{x}_{\theta,t};y_{\mathrm{tgt}},t\right)-\epsilon_{\phi}\left(\mathbf{x}_{\theta,t};\varnothing,t\right)\right)\  +\\
& \left(\epsilon_{\phi}\left(\mathbf{x}_{\theta,t};\varnothing,t\right)-\left(t<0.2\right)\cdot\epsilon_{\phi}\left(\mathbf{x}_{\theta,t};y_{\mathrm{neg}},t\right)\right)
\end{aligned}
\label{eq:nfsd}
\end{equation}
NFSD leverages the negative prompt terms to approximate domain correction, enabling noise-free score distillation.

\subsubsection{Classifier Score Distillation (CSD).}

\begin{equation}
\begin{aligned}
\delta_{\mathrm{CSD}}=&w_1\cdot\left(\epsilon_\phi\left(\mathbf{x}_{\theta,t};y_{\mathrm{tgt}},t\right)-\epsilon_\phi\left(\mathbf{x}_{\theta,t};\varnothing,t\right)\right)\ +\\
&w_2\cdot\left(\epsilon_\phi\left(\mathbf{x}_{\theta,t};\varnothing,t\right)-\epsilon_\phi\left(\mathbf{x}_{\theta,t};y_{\mathrm{neg}},t\right)\right)
\end{aligned}
\label{eq:csd}
\end{equation}
CSD gradually reduces $w_2$ mitigates the negative effects of the latter to improve texture quality, fidelity, and alignment with the target prompt.

\subsubsection{Bridge.}

\begin{equation}
\delta_{\mathrm{Bridge}}=w\cdot(\epsilon_\phi\left(\mathbf{x}_{\theta,t};y_{\mathrm{tgt}},t\right)-\epsilon_\phi\left(\mathbf{x}_{\theta,t};y_{\mathrm{tnp}},t\right))
\label{eq:bridge}
\end{equation}
Bridge improves source distribution estimation by using negative prompts to describe image corruptions.

\subsubsection{The trade-off of these variants.}
In practice, both NFSD and CSD outperform SDS, but the improvement is limited. Bridge achieves a significantly larger improvement over SDS and can generate rich, clear textures and realistic colors, though it introduces shape distortions. This reflects a clear trade-off: texture optimization either remains limited, or significant improvements in texture fidelity come at the cost of shape distortion. Although these variants analyze SDS from different perspectives and use negative prompts for optimization, they can all be transformed into the same structure. NFSD and CSD share an identical structure, to better understand the differences between Bridge, NFSD, and CSD, we reformulate Bridge into a structure consistent with the latter two, as in Equation \ref{eq:bridge2}.
\begin{equation}
\begin{aligned}
\delta_{\mathrm{Bridge}}=&w\cdot\left(\epsilon_\phi\left(\mathbf{x}_{\theta,t};y_{\mathrm{tgt}},t\right)-\epsilon_\phi\left(\mathbf{x}_{\theta,t};\varnothing,t\right)\right)\ +\\
&w\cdot\left(\epsilon_\phi\left(\mathbf{x}_{\theta,t};\varnothing,t\right)-\epsilon_\phi\left(\mathbf{x}_{\theta,t};y_{\mathrm{tnp}},t\right)\right)
\end{aligned}
\label{eq:bridge2}
\end{equation}
By comparing Equations \ref{eq:nfsd}, \ref{eq:csd}, and \ref{eq:bridge2}, our comparative experimental analysis reveals that this trade-off is fundamentally governed by the utilization of negative prompt. Details of the comparative analysis are provided in the supplementary material.  Specifically, Bridge adopts \textbf{Target negative prompts (TNP)}, which contain target information. The form of $y_{\mathrm{tnp}}$ is:
\begin{equation}
y_{\mathrm{tnp}} = y_{\mathrm{tgt}} + y_{\mathrm{neg}}
\end{equation}
For example, if $y_{\mathrm{tgt}}$ is ``An ice cream sundae'' and $y_{\mathrm{neg}}$ is ``, oversaturated, smooth, pixelated...'', then $y_{\mathrm{tnp}}$ is ``An ice cream sundae, oversaturated, smooth, pixelated...''. To avoid ambiguity, negative prompts lacking target information will be uniformly referred to as ``General Negative Prompts (GNP)'' hereafter.

\subsection{The Impact of Target Negative Prompts}
SDS optimizes 3D parameters indirectly by refining rendered images of 3D models, which is essentially an optimization of 2D images.  In order to analyze why TNP can promote the generation of clear colors while causing shape distortions, we visualize the latter term of this structure (i.e., $\delta_{\mathrm{post}}$) in Equations \ref{eq:nfsd}, \ref{eq:csd}, and \ref{eq:bridge2}, along with classifier-free guidance terms (i.e., $\delta^{\mathrm{cls}}$) and Bridge gradients (i.e., $\delta_{\mathrm{Bridge}}$). The form of $\delta_{\mathrm{post}}$ is:
\begin{equation}
\begin{aligned}
&\delta_{\mathrm{post}}^{\mathrm{gnp}}=\epsilon_\phi\left(\mathbf{x}_{\theta,t};\varnothing,t\right)-\epsilon_\phi\left(\mathbf{x}_{\theta,t};y_{\mathrm{gnp}},t\right) \\
&\delta_{\mathrm{post}}^{\mathrm{tnp}}=\epsilon_\phi\left(\mathbf{x}_{\theta,t};\varnothing,t\right)-\epsilon_\phi\left(\mathbf{x}_{\theta,t};y_{\mathrm{tnp}},t\right)
\end{aligned}
\label{eq:post}
\end{equation}

As shown in Figure \ref{fig:method2}, $\delta_{\mathrm{post}}^{\mathrm{gnp}}$ acts globally on the whole image, $\delta_{\mathrm{post}}^{\mathrm{tnp}}$ acts more focused on the target information in the image and can more accurately suppress negative states in the target region, which is referred to as the \textbf{focused suppression effect}. Furthermore, a comparison between $\delta_{\mathrm{Bridge}}$ and $\delta^{\mathrm{cls}}$ reveals that the target information in $\delta_{\mathrm{Bridge}}$ is relatively vague, which reflects the \textbf{target information loss} caused by TNP. This loss impairs the concentration of gradients on target regions, thereby leading to shape distortion.

\subsection{Analysis of these Variants}
From the perspective of negative prompts, NFSD and CSD employ GNP, whose suppression of negative states fails to focus on the target itself, limiting texture improvement. In contrast, Bridge uses TNP, which accurately focuses on and suppresses negative states in target regions to effectively optimize textures but causes target information loss, thereby leading to shape distortion. To achieve generated results with clear textures and stable shapes, we need to leverage TNP for texture optimization while enhancing protection of target information.

\begin{figure*}[t!]  
\centering
\includegraphics[width=1\textwidth]{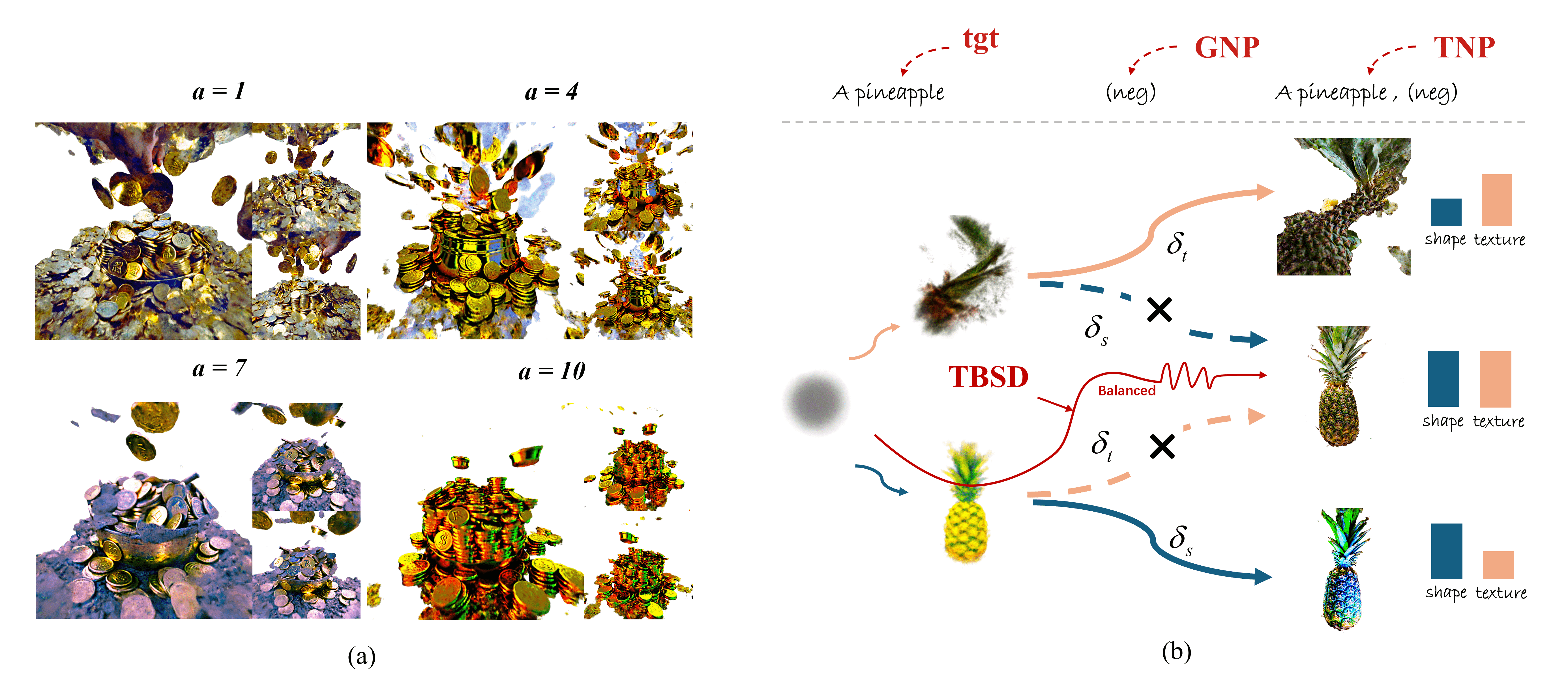} 
\caption{(a) Result images with varying levels of shape information controlled by coefficient \( a \). (b) An overview of the proposed TBSD. Solid lines indicate reachable optimization paths, while dashed lines indicate unreachable ones. The prompts used for Figs. (a), (b) are "A cauldron full of gold coins", "A pineapple", respectively. } 
\label{fig:tbsd} 
\end{figure*}

\section{Method}
From the aforementioned analysis, TNP’s core focusing suppression effect enhances the realism and detail clarity of texture generation by accurately suppressing negative states in the target region. However, it also leads to shape distortion of the generated object due to the loss of target information.

To address this issue, an intuitive solution is to inject target information into the optimization gradient to supplement geometric constraints. The classifier-free guidance term in SDS maximizes the difference in target information through the gap between the target prompt and empty prompt, thereby effectively providing target information. Based on this, we introduce this guidance term into the optimization framework of TNP and control the injection intensity of target information by adjusting its coefficient \(a\) (see Equation \ref{eq:finalv1}).
\begin{equation}
\begin{aligned}
\delta_x(\mathbf{x}_t;y,t)  = &\epsilon_\phi\left(\mathbf{x}_{\theta,t};y_{\mathrm{tgt}},t\right)-\epsilon_\phi\left(\mathbf{x}_{\theta,t};y_{\mathrm{tnp}},t\right)\ \\ \ &+ a\cdot\delta^{\mathrm{cls}}  
\label{eq:finalv1}
\end{aligned}
\end{equation}

We inject target information of varying intensities into the TNP optimization process. As shown in Figure \ref{fig:tbsd} (a), when the value of $a$ is small, the generated results exhibit clear textures but distorted shapes. when $a$ is too large, the shape becomes accurate but the texture appears overly saturated in color. This indicates that a fixed coefficient $a$ fails to achieve a proper balance between texture and shape.

To address this problem, we propose \textbf{Target-Balanced Score Distillation (TBSD)}, inspired by Multiple-Gradient Descent Algorithm (MGDA) \cite{mgda}. TBSD formulates the generation task as a multi-objective optimization problem, with shape ($\delta_{s}$) and texture ($\delta_{t}$) as the two objectives. Since the classifier-free guidance term provides sufficient target information and TNP, with its focus-suppression effect, enables realistic and vivid texture optimization, the optimization objective is defined as in Equation \ref{eq:target}.

\begin{equation}
\begin{aligned}
 &\delta_{s}  =  \epsilon_\phi\left(\mathbf{x}_{\theta,t};y_{\mathrm{tgt}},t\right)-\epsilon_\phi\left(\mathbf{x}_{\theta,t};\varnothing,t\right) \\
 &\delta_{t}   =  \epsilon_\phi\left(\mathbf{x}_{\theta,t};y_{\mathrm{tgt}},t\right)-\epsilon_\phi\left(\mathbf{x}_{\theta,t};y_{\mathrm{tnp}},t\right)
\end{aligned}
\label{eq:target}
\end{equation}

The cosine similarity between the texture and shape objectives is always positive, indicating that they share a similar optimization direction. In this case, MGDA tends to favor the objective with a smaller gradient norm if the other is significantly larger. Based on this property and the progressive nature of 3D generation, we observe that a good initial shape helps guide later optimization and improves final quality. Therefore, we introduce a dynamic weighting factor for the texture objective, as shown in Equation \ref{eq:factor}. 
\begin{equation}
\begin{aligned}
&\delta_{td}  =factor(t)  \cdot \delta_{t}   \\
&factor(t) = max(\alpha \cdot (1-t/\beta),\gamma)
\end{aligned}
\label{eq:factor}
\end{equation}
This factor starts with a large value $\alpha$, decreases over time with rate $\frac{\alpha}{\beta}$, and is bounded below by $\gamma$. $t$ is iteration step.

The factor ensures that TBSD initially focuses more on shape optimization, enabling the model to generate an accurate shape in the early stages. As training progresses, the optimization gradually shifts toward the texture objective. During this process, TBSD maintains a good balance between shape and texture, allowing texture quality to improve while preserving essential target information. Eventually, it reaches an optimal optimization point for both shape and texture. Based on this, the formula of TBSD is shown in Equation \ref{eq:TBSD}.

\begin{equation}
\begin{aligned}
\delta_x^\mathrm{TBSD}=\mu \cdot \delta_{s}+(1-\mu)\cdot \delta_{td}
\end{aligned}
\label{eq:TBSD}
\end{equation}
Following \cite{mgda-mtl}, the multi-objectives optimization problem can be defined as:
\begin{equation}
\min_{\mu,1-\mu\in[0,1]}\|\mu \cdot \delta_{s}+(1-\mu)\cdot\delta_{td})\|_2^2
\end{equation}
This is a univariate quadratic function of $\mu$, and considering the value range of $\mu$, its solution is:
\begin{equation}
\begin{aligned}
\mu=\min\left[\max\left[0,\frac{(\delta_{td}-\delta_{s})^T\delta_{td}}{\|\delta_{td}-\delta_{s}\|_2^2}\right],1\right]
\end{aligned}
\label{eq:mu}
\end{equation}

As a result, TBSD effectively resolves the aforementioned trade-off and can achieve the generation of 3D objects with accurate shapes and high-fidelity textures. Moreover, TBSD can be directly applied to image generation. TBSD is shown in Figure \ref{fig:tbsd} (b).

\section{Experiments}
We rigorously evaluate the proposed method on both 2D and 3D generation tasks employing Score Distillation Sampling (SDS), conducting comprehensive benchmarks against SDS and domain-specific state-of-the-art baselines. Extended quantitative analyses, qualitative comparisons, and ablation studies are detailed in the supplementary material.
\subsection{Implementation Details}

We implement text-based 3D generation using the threestudio \cite{threestudio} framework. Unless specified otherwise, all 3D models are trained with the AdamW \cite{adamw} optimizer for 20,000 iterations, with a learning rate of 0.01. Our initial rendering resolution is gradually increased from 64×64 to 256×256. Consistent with NFSD \cite{NFSD} and Bridge \cite{Bridge}, implicit volumes use an object-centered initialization strategy \cite{magic123,ProlificDreamer}. All experiments employ the pre-trained text-to-image diffusion model Stable Diffusion 2.1-base \cite{t2i}. Additional implementation details are provided in the supplementary material.

\begin{figure*}[t!]  
\centering
\includegraphics[width=\textwidth]{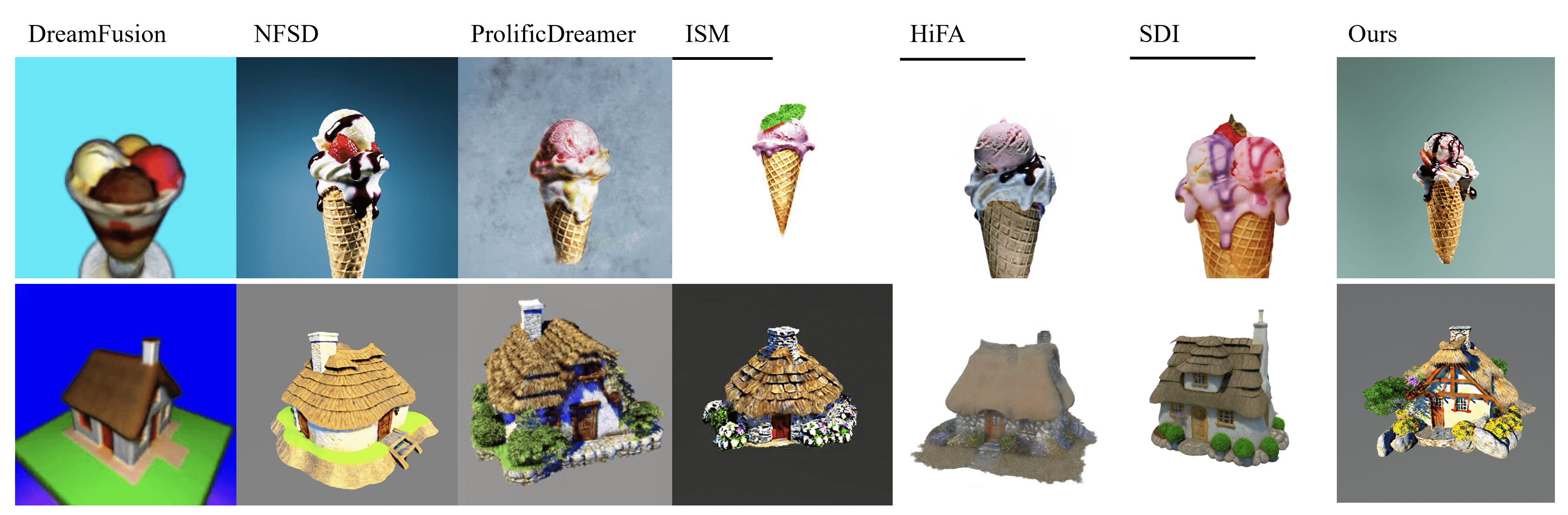} 
\caption{3D generation comparison with other methods, using their reported results. The prompts employed are “An ice cream sundae” and “A 3D model of an adorable cottage with a thatched roof”.} 
\label{fig:baseline} 
\end{figure*}

\subsection{Text-to-3D Generation}

\subsubsection{Qualitative comparisons.}

Our comparison with recent methods is presented in Figure \ref{fig:baseline}. Following the same experimental protocol as NFSD \cite{NFSD}, ProlificDreamer \cite{ProlificDreamer}, and SDI \cite{SDI}, we compare the 3D generation quality of this study with previous works. Selected baseline methods include DreamFusion \cite{DreamFusion}, NFSD \cite{NFSD}, ProlificDreamer \cite{ProlificDreamer}, ISM \cite{LucidDreamer}, HiFA \cite{HIFA}, and SDI \cite{SDI}, with all comparison results from the original authors. It can be observed that our method achieves comparable or better results without additional computation from model training or multi-step optimization. More comparisons of text-to-3D experimental results are available in supplementary material.

\subsubsection{Quantitative Results.}

We perform quantitative evaluation of generation quality following \cite{DreamFusion,CSD,SDI}. The Clip scores in Table \ref{tab:data_table} are computed using torchmetrics \cite{torchmetrics} and the ViT-B/32 model \cite{vitb32}. We test 50 views under 43 prompts \cite{SDI}. For fairness, multi-stage methods only run the first stage. All results are compared based on threestudio reproductions. It can be seen that TBSD outperforms SDS in quality and the current state-of-the-art method SDI \cite{SDI}, and notably, this improvement is achieved without multi-step optimization. To further assess the perceptual quality of generated results, we conduct a user study comparing our approach with baselines. As shown in Table \ref{tab:data_table}, our method outperforms the baselines in the user study. Details of the user study are presented in the supplementary material.

\begin{table}[t]
    \centering
    \setlength{\tabcolsep}{3pt}
    \renewcommand{\arraystretch}{1.2}
    \begin{tabular}{c c c}
    \toprule
    Method & CLIP Score($ \uparrow $) & User Preference (\%) ($ \uparrow $) \\
    \midrule

    SDS   & 29.81 & 2.61 \\
    VSD   & 33.31 & 11.52 \\
    HIFA  & 32.80 & 7.48\\
    SDI  & 33.47 & 13.44 \\
    \hline
    CSD  & 32.05 & 7.73 \\
    NFSD & 31.89 & 6.56 \\
    Bridge  & 32.36 & 8.29 \\
    \textbf{TBSD} & \textbf{33.59} & \textbf{42.37} \\
    \bottomrule
    \end{tabular}
    \caption{The quantitative comparison of 3D generation between our method and others.}
    \label{tab:data_table}
\end{table}

\subsection{Ablation Studies and Analysis}

\begin{figure}[t]
  \includegraphics[width=0.47\textwidth]{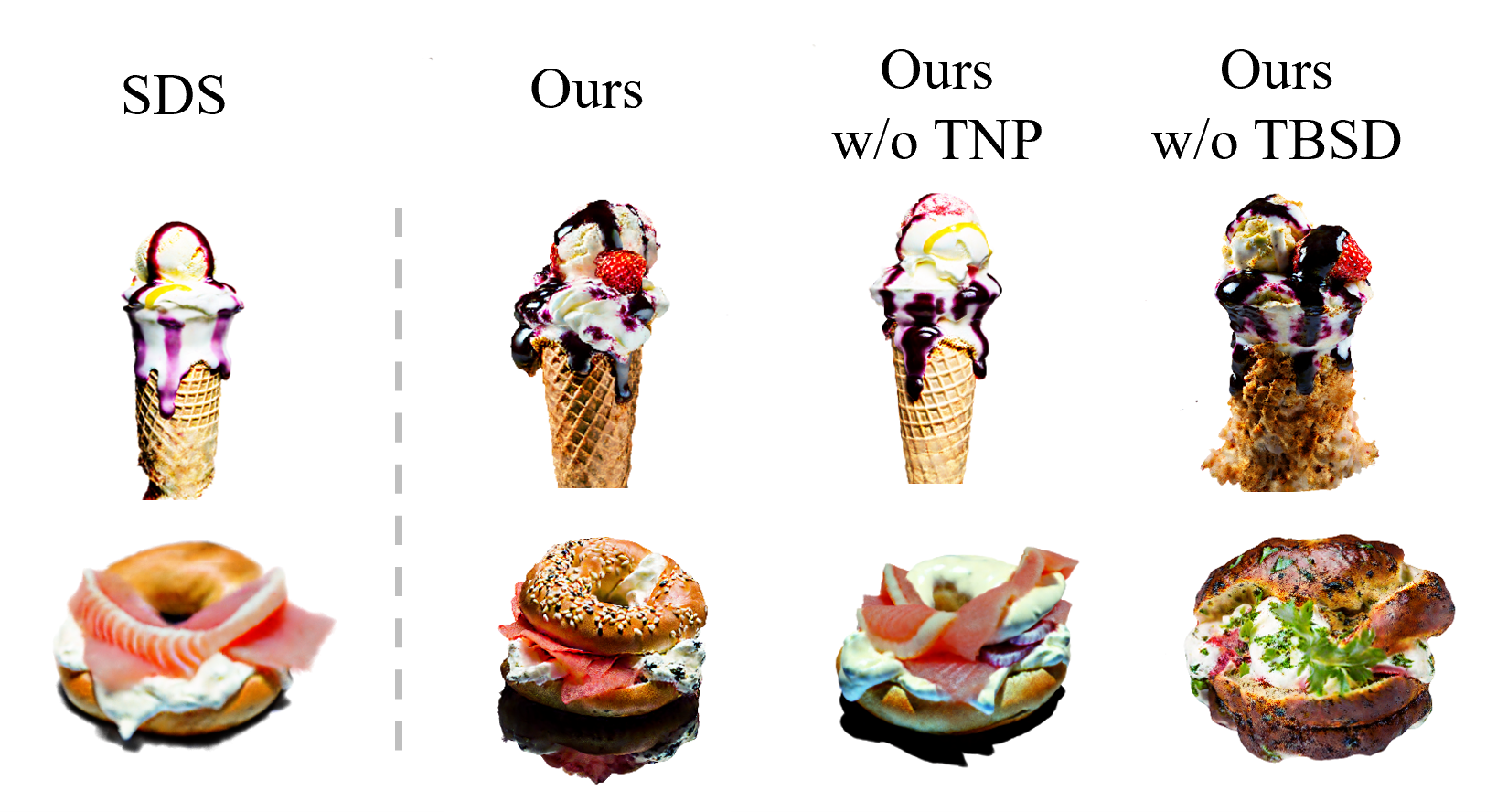} 
  \caption{Ablation study of proposed improvements. First and second row results use prompts “An ice cream sundae” and “Bagel filled with cream cheese and lox”, respectively.} 
  \label{fig:Ablation study of proposed improvements} 
\end{figure}

\subsubsection{Ablation study of proposed improvements.}

Figure \ref{fig:Ablation study of proposed improvements} shows ablation study of our proposed improvements. It can be observed that using TNP significantly improves generation quality, resulting in richer textures and more realistic, vivid colors. Moreover, TBSD effectively resolves the trade-off between shape and texture, ultimately producing 3D outputs with accurate shape and high-fidelity, realistic textures.

\subsubsection{Negative Prompts Ablation.}
\begin{figure}[t!]
  \centering
  \includegraphics[width=0.47\textwidth]{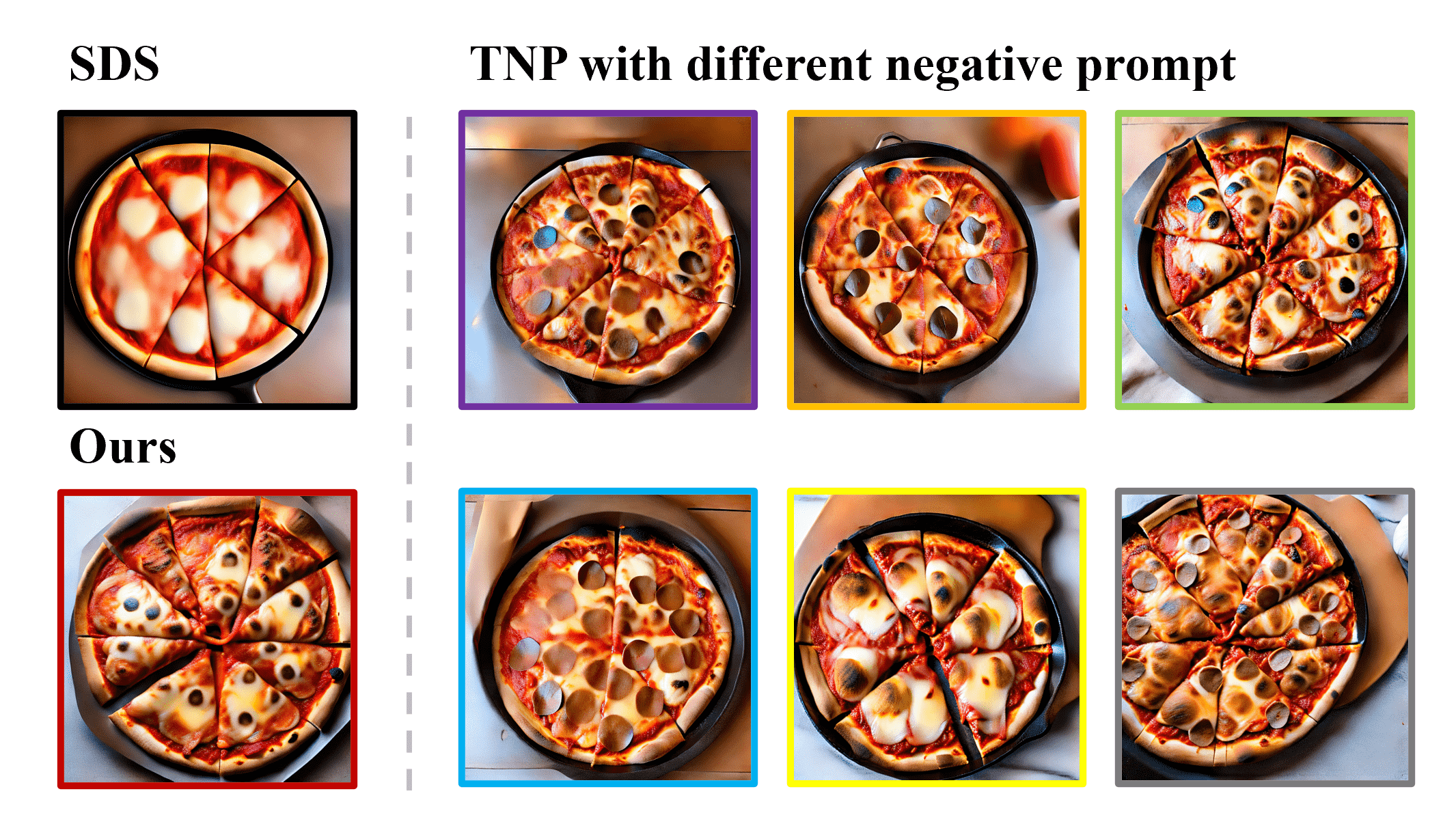} 
  \caption{Ablation study on negative prompts in TNP. These images are generated by TBSD using TNP with varying negative prompts. The prompt is “pizza sitting on top of a pan on a table”.} 
  \label{fig:Negative Prompt Ablation} 
\end{figure}

We explore the impact of negative prompts in TNP on generation quality. We generate six groups of different negative prompts descriptions via GPT-4 for comparative evaluation. In the experiment, all other hyperparameters are kept unchanged, with only the negative text following the target text replaced. Specific results are shown in Figure \ref{fig:Negative Prompt Ablation}. No significant differences are observed across the ablation experiments with various negative prompts in TNP, indicating that the design of the TNP prompt format is more critical than the specific wording of negative prompts used. Details of negative prompt variants are provided in supplementary material.

\subsubsection{Hyperparameter Ablation.}

\begin{figure}[t!]
\centering
  \includegraphics[width=0.46\textwidth]{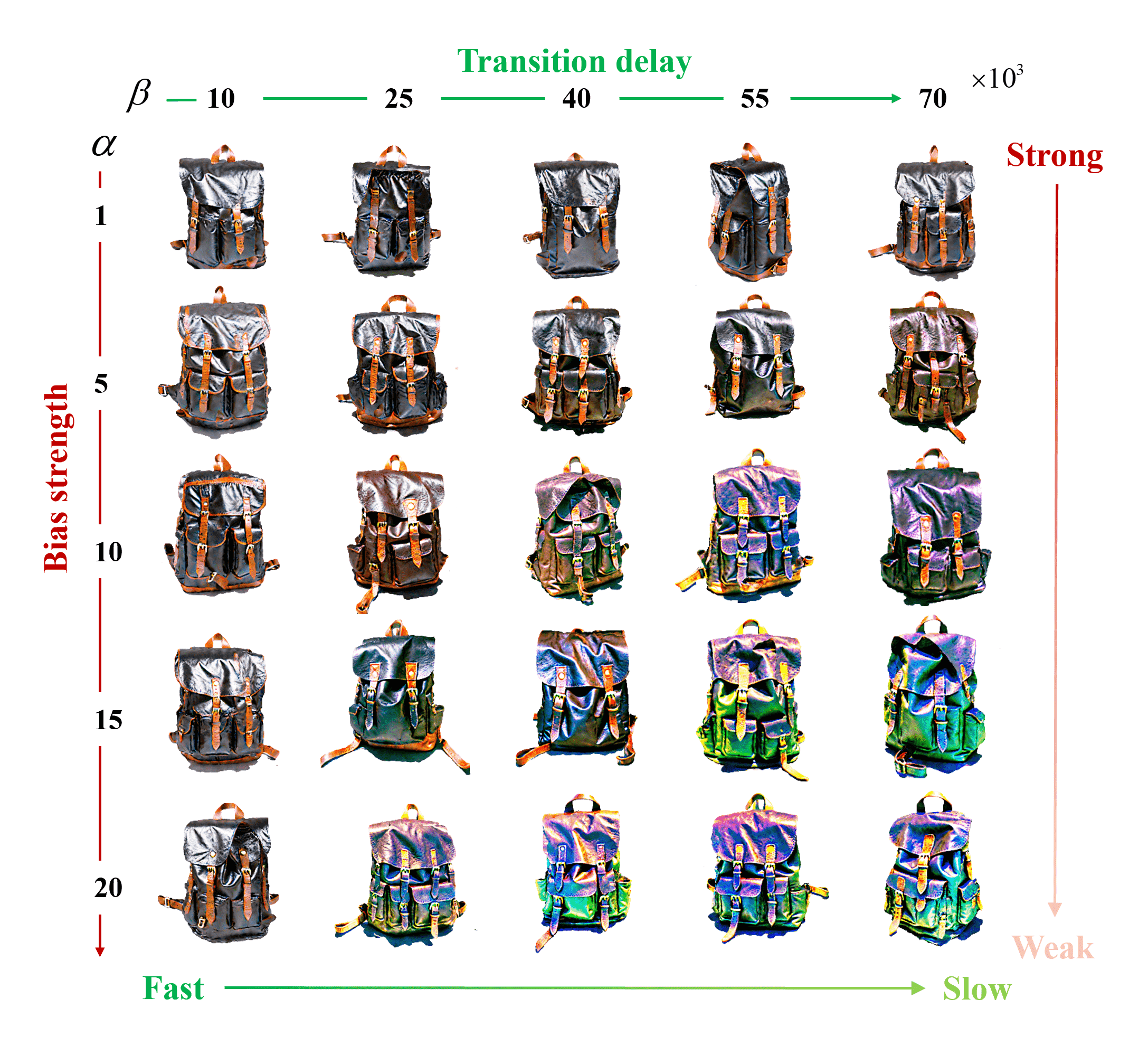} 
  \caption{Ablation result of the $\alpha$ and $\beta$. The prompt is ``Photograph of a black leather backpack".} 
  \label{fig:Hyperparameter Ablation} 
\end{figure}

We investigate the impact of two tunable parameters in Equation \ref{eq:factor} : $\alpha$ and $\beta$. 
The parameter $\alpha$ controls the strength of TBSD's focus bias toward shape optimization, while \(\beta\) regulates the delay in shifting this focus toward texture. As shown in Figure \ref{fig:Hyperparameter Ablation}, large values for both $\alpha$ and $\beta$ result in a strong and persistent shape bias, with a slow transition toward texture optimization. This leads to generated 3D assets with oversaturated colors and insufficient texture details. Only with a proper combination of $\alpha$ and $\beta$ can TBSD gradually and effectively shift its optimization focus from shape to texture, leading to realistic 3D outputs with sharp and detailed textures.

\subsection{Text-to-img Generation}

\begin{figure}[t!]
\centering
  \includegraphics[width=0.47\textwidth]{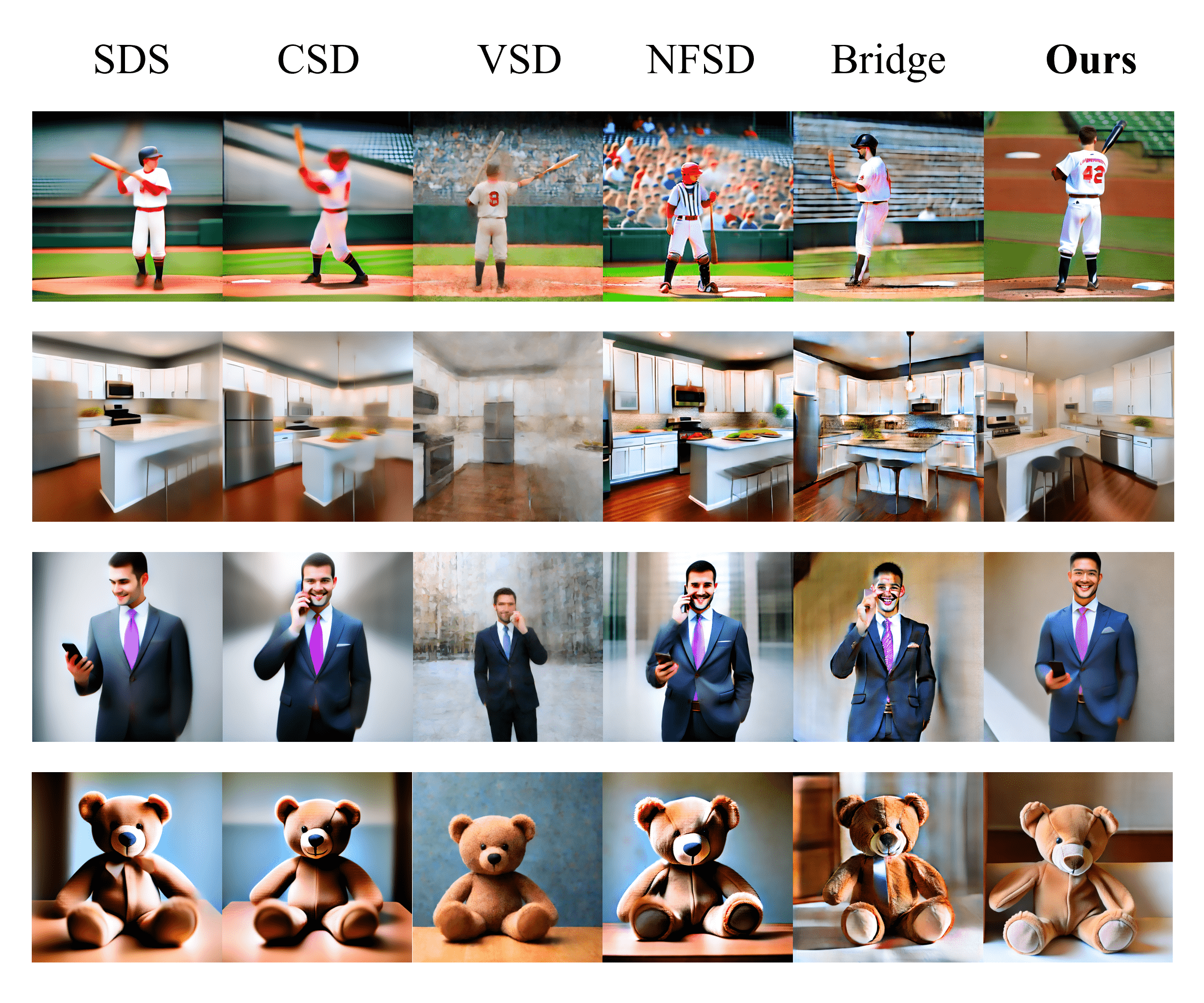} 
  \caption{Text-to-image generation results using COCO Captions. We compare various score distillation methods for image generation with COCO captions, where images are optimized from random initializations.} 
  \label{fig:2d比较} 
\end{figure}

To verify our analysis of existing SDS variants and the proposed method, we also conduct text-to-image generation experiments by optimizing images in the Stable Diffusion latent space consistent with previous studies. Compared with text-to-3D tasks, where factors such as initialization strategies, 3D representation methods, and 2D prior models can significantly affect the final results, image generation is less influenced by such confounding variables. To illustrate the advantages of TBSD in 2D experiments, we selected some MS-COCO prompts \cite{coco} for comparative display. Comparative methods include SDS, VSD, as well as NFSD, CSD, and Bridge that take negative prompts as the core. For each prompt, we randomly initialize the noise and then optimize using score distillation gradients. Figure \ref{fig:2d比较} shows generation examples of different score distillation methods, SDS and CSD exhibit both over-saturation and excessive smoothness. NFSD shows an improved texture, but still tends to be saturated. VSD and Bridge generate samples that are closest to real-world effects, but VSD suffers from global blurriness. Although Bridge exhibits relatively rich details, it has slight saturation issues and shape distortion. In contrast, our method achieves realistic colors, rich textures, and stable shapes, thereby outperforming these approaches. More details of 2D experiments can be seen in the supplementary material.

\section{Conclusion}
In this paper, we first systematically analyze the trade-off between texture fidelity and shape accuracy in SDS methods utilizing negative prompts and reveal that Target Negative Prompts (TNP) enhance texture realism by suppressing negative states in the target region, but this strong focus can lead to the loss of target information, resulting in shape distortions. To address this, we introduce Target-Balanced Score Distillation (TBSD), a novel multi-objective framework that adaptively balances shape and texture optimization by progressively shifting focus from global shape to detailed texture refinement. Extensive experiments demonstrate that TBSD effectively resolves the trade-off, producing 3D assets with both accurate shape and rich textures, outperforming existing approaches. We believe that our findings can offer a novel insight for the SDS community.

\section{Acknowledgments}
This work is supported by the Shenzhen Science and Technology Project under Grant KJZD20240903103210014.

\bibliography{aaai2026}

\newpage

\setlength\columnsep{0.375in}
\setlength\titlebox{2.25in}

\twocolumn[{
    \centering
    \vspace{0.625in}
    {\LARGE\bf Supplementary Material \par} 
    \vspace{1cm}
}]

\setcounter{section}{0}
\setcounter{figure}{0}
\setcounter{table}{0}
\setcounter{equation}{0}

\begin{figure*}[ht!]
    \centering
  \includegraphics[width=1\textwidth]{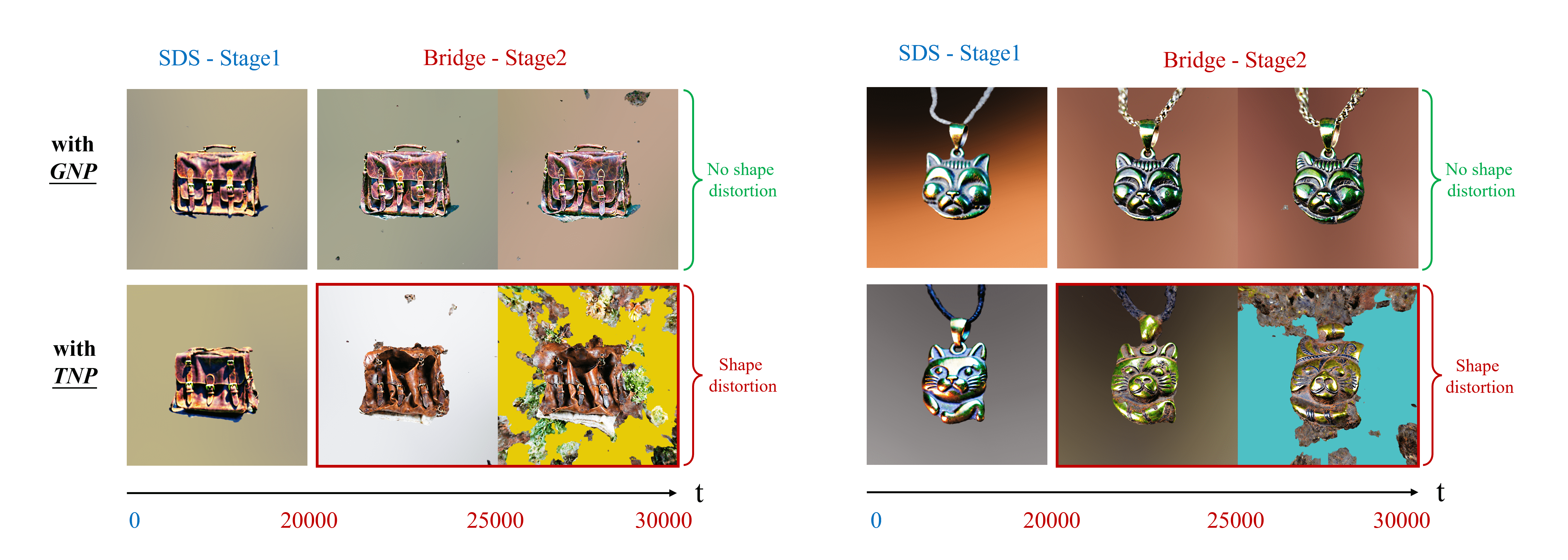} 
  \caption{Comparison results of Bridge with TNP and GNP. For images corresponding to the two sets of prompts, the leftmost column shows results generated by SDS in the first stage of Bridge, the right two columns show results optimized over time, and the prompts used are "A worn-out leather briefcase" and "A tumbaga pendant depicting a cat" from left to right.} 
  \label{fig:analysis} 
\end{figure*}

This supplementary material provides comprehensive additional analyses and experimental details to support the findings presented in the main paper. We begin with a comparative analysis of SDS-based methods using different negative prompts, illustrating the effects of Target Negative Prompts (TNP) and Generic Negative Prompts (GNP) on generation quality and shape preservation. We then detail the derivation and interpretation of our proposed Target-Balanced Score Distillation (TBSD) formulation. Extensive implementation details are provided for both text-to-image and text-to-3D tasks, including all hyperparameter settings and optimization procedures. A user study involving 65 participants across 35 prompts is described, highlighting the clear preference for our method. Additional ablation studies explore the role of negative prompts and output diversity. We further present a wide range of generation results and qualitative comparisons with state-of-the-art baselines, demonstrating the robustness, fidelity, and generalizability of our approach. Lastly, we provide a full list of prompts used in the qualitative evaluations for reproducibility.

\section{Additional Analysis on Negative Prompts}
This section presents an extended analysis of negative prompts, highlighting that the incorporation of negative textual guidance is a primary factor underlying the trade-offs observed in several existing methods.

Building upon the analysis presented in Section 2 of the main paper on negative-prompt-based methods—namely Noise-Free Score Distillation (NFSD), Classifier Score Distillation (CSD) , and Bridge—we revisit their respective formulations, which are structured as follows:
\begin{equation}
\begin{aligned}
\delta_{\mathrm{NFSD}}=\  &s\cdot\left(\epsilon_{\phi}\left(\mathbf{x}_{\theta,t};y_{\mathrm{tgt}},t\right)-\epsilon_{\phi}\left(\mathbf{x}_{\theta,t};\varnothing,t\right)\right)\  +\\
& \left(\epsilon_{\phi}\left(\mathbf{x}_{\theta,t};\varnothing,t\right)-\left(t<0.2\right)\cdot\epsilon_{\phi}\left(\mathbf{x}_{\theta,t};y_{\mathrm{gnp}},t\right)\right) \\
\delta_{\mathrm{CSD}}=\ &w_1\cdot\left(\epsilon_\phi\left(\mathbf{x}_{\theta,t};y_{\mathrm{tgt}},t\right)-\epsilon_\phi\left(\mathbf{x}_{\theta,t};\varnothing,t\right)\right)\ +\\
&w_2\cdot\left(\epsilon_\phi\left(\mathbf{x}_{\theta,t};\varnothing,t\right)-\epsilon_\phi\left(\mathbf{x}_{\theta,t};y_{\mathrm{gnp}},t\right)\right) \\
\delta_{\mathrm{Bridge}}=\ &w\cdot\left(\epsilon_\phi\left(\mathbf{x}_{\theta,t};y_{\mathrm{tgt}},t\right)-\epsilon_\phi\left(\mathbf{x}_{\theta,t};\varnothing,t\right)\right)\ +\\
&w\cdot\left(\epsilon_\phi\left(\mathbf{x}_{\theta,t};\varnothing,t\right)-\epsilon_\phi\left(\mathbf{x}_{\theta,t};y_{\mathrm{tnp}},t\right)\right)
\end{aligned}
\label{eq:same}
\end{equation}
As discussed in the main text, Generic Negative Prompts (GNP) do not contain any target-related information, whereas Target Negative Prompts (TNP) explicitly encode aspects of the target object.

From Equation \ref{eq:same}, the most intuitive distinction among the methods lies in the design of their negative prompts. To investigate whether performance differences arise from this variation, we conduct a direct substitution: replacing TNP with GNP in the Bridge framework and comparing the results before and after the replacement.

As illustrated in Figure \ref{fig:analysis}, employing TNP in Bridge significantly improves texture quality, but at the cost of noticeable shape distortion. In contrast, using GNP leads to only marginal texture enhancement. These observations suggest that the trade-off between texture fidelity and shape consistency is fundamentally driven by the nature of the negative prompts used.

\section{Relationship with Classifier-Free Guidance(CFG)}
\subsection{Link to CFG}
SDS uses a large-weight CFG coefficient to let the classification term dominate 3D generation, $\delta_{\mathrm{SDS}}$ with CFG is expressed as:
\begin{equation}
\begin{aligned}
\delta_{\mathrm{SDS}}=& \epsilon_{\phi}\left(\mathbf{x}_{\theta,t};\varnothing,t\right)-\epsilon \ +\\
& s\cdot \underbrace{\left(\epsilon_{\phi}\left(\mathbf{x}_{\theta,t};y_{\mathrm{tgt}},t\right)-\epsilon_{\phi}\left(\mathbf{x}_{\theta,t};\varnothing,t\right)\right)}_{\delta^{\mathrm{cls}}}
\end{aligned}
\label{eq:sds2}
\end{equation}
Since CFG is a commonly compared factor in SDS, we convert TBSD into the same form as SDS to intuitively observe the CFG coefficient. The converted form of TBSD is as follows: 
\begin{equation}
\begin{aligned}
\delta_{\mathrm{TBSD}}=&\epsilon_\phi\left(\mathbf{x}_{\theta,t};\varnothing,t\right)-\epsilon_\phi\left(\mathbf{x}_{\theta,t};y_{\mathrm{tnp}},t\right)\ +\\
&\frac{1}{1-\mu}   \cdot\left(\epsilon_\phi\left(\mathbf{x}_{\theta,t};y_{\mathrm{tgt}},t\right)-\epsilon_\phi\left(\mathbf{x}_{\theta,t};\varnothing,t\right)\right)
\end{aligned}
\label{eq:cfg}
\end{equation}

Comparing Eq. \ref{eq:sds2} and Eq. \ref{eq:cfg}, the CFG coefficient in Eq. \ref{eq:cfg} is $\frac{1}{1-\mu}$ where $\mu$ is calculated by TBSD, meaning the CFG coefficient of our method dynamically changes with 3D states to regulate TNP impacts, ultimately achieving stable shapes and clear textures. 

\subsection{Detailed TBSD conversion process}
To more clearly illustrate the structure of CFG in TBSD, we set the factor (as defined in Equation 11 of the main text) to 1. In this case, the TBSD formula can be expressed as:
\begin{equation}
\begin{aligned}
\delta_{\mathrm{TBSD}}=&   \mu \cdot  \left(\epsilon_\phi\left(\mathbf{x}_{\theta,t};y_{\mathrm{tgt}},t\right)\  - \epsilon_\phi\left(\mathbf{x}_{\theta,t};\varnothing,t\right)\right) \ +\\
&\ (1-\mu)   \cdot \left(\epsilon_\phi\left(\mathbf{x}_{\theta,t};y_{\mathrm{tgt}},t\right)-\epsilon_\phi\left(\mathbf{x}_{\theta,t};y_{\mathrm{tnp}},t\right)\right)
\end{aligned}
\end{equation}
The detailed conversion process is as follows:
\begin{equation}
\begin{aligned}
\delta_{\mathrm{TBSD}}=& \  \mu \cdot  (\epsilon_\phi\left(\mathbf{x}_{\theta,t};y_{\mathrm{tgt}},t\right)\  - \epsilon_\phi\left(\mathbf{x}_{\theta,t};\varnothing,t\right)) \ +\\
&\ (1-\mu)   \cdot \left(\epsilon_\phi\left(\mathbf{x}_{\theta,t};y_{\mathrm{tgt}},t\right)-\epsilon_\phi\left(\mathbf{x}_{\theta,t};\varnothing,t\right)\right) \ + \\
&\ (1-\mu)     \cdot  \left(\epsilon_\phi\left(\mathbf{x}_{\theta,t};\varnothing,t\right) - \epsilon_\phi\left(\mathbf{x}_{\theta,t};y_{\mathrm{tnp}},t\right)\right) \\
= & \ (1-\mu)     \cdot  \left(\epsilon_\phi\left(\mathbf{x}_{\theta,t};\varnothing,t\right) - \epsilon_\phi\left(\mathbf{x}_{\theta,t};y_{\mathrm{tnp}},t\right)\right)\ +\\ 
 &\left(\epsilon_\phi\left(\mathbf{x}_{\theta,t};y_{\mathrm{tgt}},t\right)\  - \epsilon_\phi\left(\mathbf{x}_{\theta,t};\varnothing,t\right)\right) \\
 = & \ (1-\mu)    \cdot    [\epsilon_\phi\left(\mathbf{x}_{\theta,t};\varnothing,t\right) - \epsilon_\phi\left(\mathbf{x}_{\theta,t};y_{\mathrm{tnp}},t\right)+ \\
& \frac{1}{1-\mu} \cdot  \left(\epsilon_\phi\left(\mathbf{x}_{\theta,t};y_{\mathrm{tgt}},t\right)\  - \epsilon_\phi\left(\mathbf{x}_{\theta,t};\varnothing,t\right)\right) ]
\end{aligned}
\end{equation}
When ignoring the influence of weights, the TBSD expression becomes:
\begin{equation}
\begin{aligned}
\delta_{\mathrm{TBSD}}=&\epsilon_\phi\left(\mathbf{x}_{\theta,t};\varnothing,t\right)-\epsilon_\phi\left(\mathbf{x}_{\theta,t};y_{\mathrm{tnp}},t\right)\ +\\
&\frac{1}{1-\mu}   \cdot\left(\epsilon_\phi\left(\mathbf{x}_{\theta,t};y_{\mathrm{tgt}},t\right)-\epsilon_\phi\left(\mathbf{x}_{\theta,t};\varnothing,t\right)\right)
\end{aligned}
\end{equation}
This reformulation intuitively separates the influence of texture-focused guidance from global shape optimization, allowing TBSD to adaptively balance the two objectives.

\section{Implementation Details}
Our experiments default to the stable-diffusion-v2-1-base model, and all were conducted on an NVIDIA 3090 GPU.

\subsection{Text-to-image generation}
 For CSD, we follow the original setup with initial values \(w_1 = w_2 = 40\), and anneal \(w_2\) to 0 within the first 500 steps.
Consistent with best practices, we set \(s = 100\) for SDS, and \(s = 7.5\) for both NFSD and VSD. For Bridge, we use \(s = 40\) and \(w = 25\). Optimization starts with the \(\epsilon_{\text{SDS}}\) loss for 500 iterations and then switches to \(\epsilon_{\text{bridge}}\) for the subsequent 500 iterations. Our method employs parameters \(\alpha = 5\), \(\beta = 2,000\), and \(\gamma = 2\), with the default negative prompts in TNP set to ``oversaturated, smooth, pixelated, cartoon, foggy, hazy, blurry, bad structure, noisy, malformed''. All methods use a learning rate of 0.01 with 1000 iterations. For LoRA training in VSD, the learning rate is set to $1e-4$. 

\subsection{Text-to-3D generation}
For Classifier Score Distillation (CSD), we follow the original configuration, initializing the weights with $w_1 = w_2 = 1$, and gradually annealing $w_2$ to zero during optimization. In line with common practice, the guidance scale is set to $s = 100$ for Score Distillation Sampling (SDS), and $s = 7.5$ for both Noise-Free Score Distillation (NFSD) and Variational Score Distillation (VSD). For Bridge, we begin optimization with the standard SDS loss$\epsilon_{\text{SDS}}$ for 20,000 iterations, followed by an additional 5,000 iterations using the Bridge-specific loss $\epsilon_{\text{bridge}}$. For our method, we set $\alpha = 5$, $\beta = 25,000$, and $\gamma = 2$. The negative prompts used in the TNP modules of both Bridge and our approach are identical to those used in the 2D experiments to ensure consistency across modalities.

\section{Details of user study}
To enable a more comprehensive evaluation, we conducted a user study. Specifically, we selected 35 prompts and prepared results from five state-of-the-art text-to-3D methods for each prompt, with each method presented in three different views. All methods were evaluated according to the standardized benchmark provided by ThreeStudio. To ensure the robustness of the results, we randomly recruited 65 participants and asked them to select the result they perceived as having the best generation quality. The user preferences were then aggregated on the basis of these selections. Our method achieved a significant advantage in the user study, clearly demonstrating the superiority of our generated results.

\begin{figure*}[ht!]
  \includegraphics[width=1\textwidth]{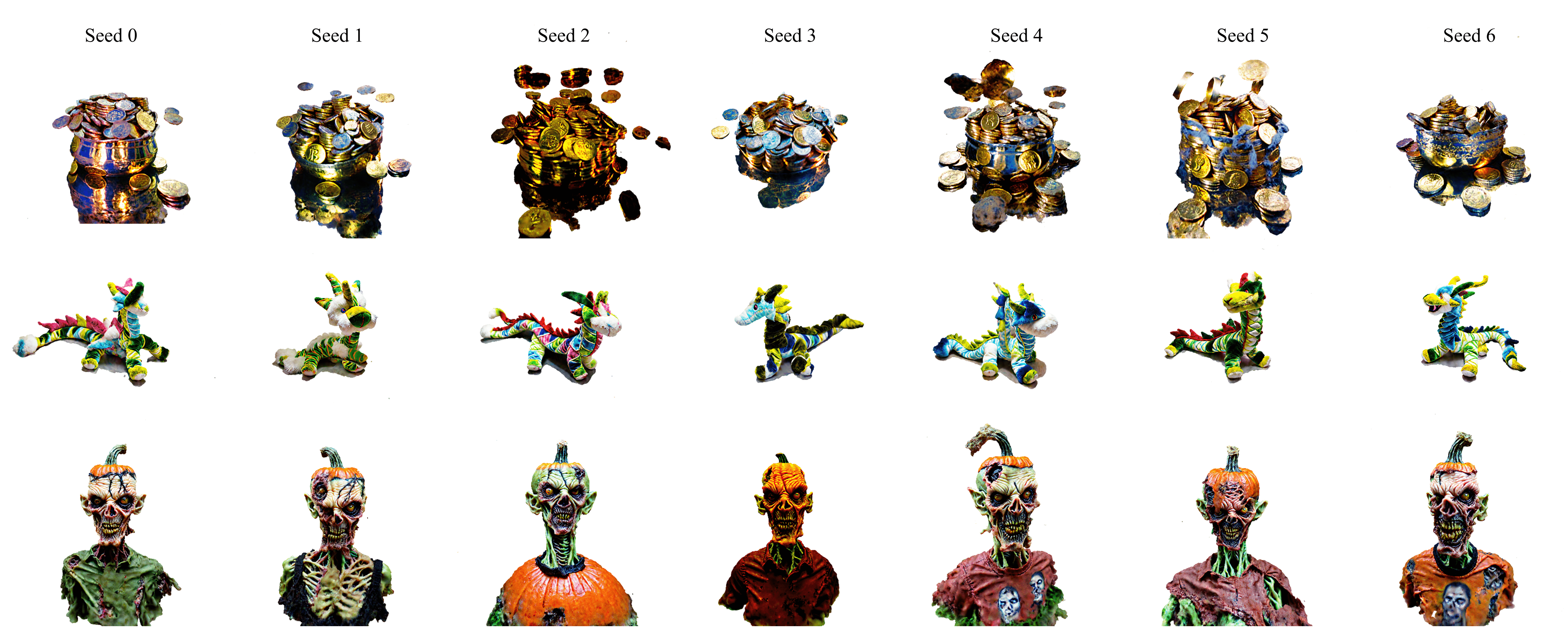} 
  \caption{Generation examples of TBSD (ours) with different seed values. Prompts are as follows: “A cauldron full of gold coins” (top row), “A plush dragon toy” (middle row), and “Pumpkin head zombie, skinny, highly detailed, photorealistic” (bottom row).} 
  \label{fig:seed} 
\end{figure*}

\section{Additional ablation Studies}

\subsection{Negative Prompts Ablation.}
The specific negative prompts used in the TNP for the "Negative Prompts Ablation" study presented in the main text are as follows:
\begin{itemize}
    \item distorted, low-res, washed-out, flat, grainy, uneven lighting, blurry edges, messy composition, muted, misshapen.
    \item smudged, oversaturated, pixelated, cartoonish, foggy, hazy, out-of-focus, poor framing, noisy, deformed, unbalanced.
    \item faded, smooth, low-contrast, pixelated, cartoon, foggy, blurry, bad proportions, grainy, malformed, discolored.
    \item oversaturated, flat, pixelated, animated, hazy, blurry, messy structure, noisy, warped, dull, misshapen.
    \item smudged, low-res, washed-out, smooth, cartoon, foggy, out-of-focus, bad composition, grainy, deformed, asymmetric.
    \item faded, oversaturated, pixelated, flat, foggy, hazy, blurry edges, poor lighting, noisy, malformed, unbalanced.
\end{itemize}

\subsection{Diversity}
Figure \ref{fig:seed} presents the results generated under different random seeds for the same prompt. Our method maintains overall shape consistency while exhibiting noticeable diversity in fine details, which can be particularly beneficial in practical applications.
 
\section{Additional Qualitative Results}

\subsection{Additional generations}
\begin{figure*}[ht!]
  \includegraphics[width=1\textwidth]{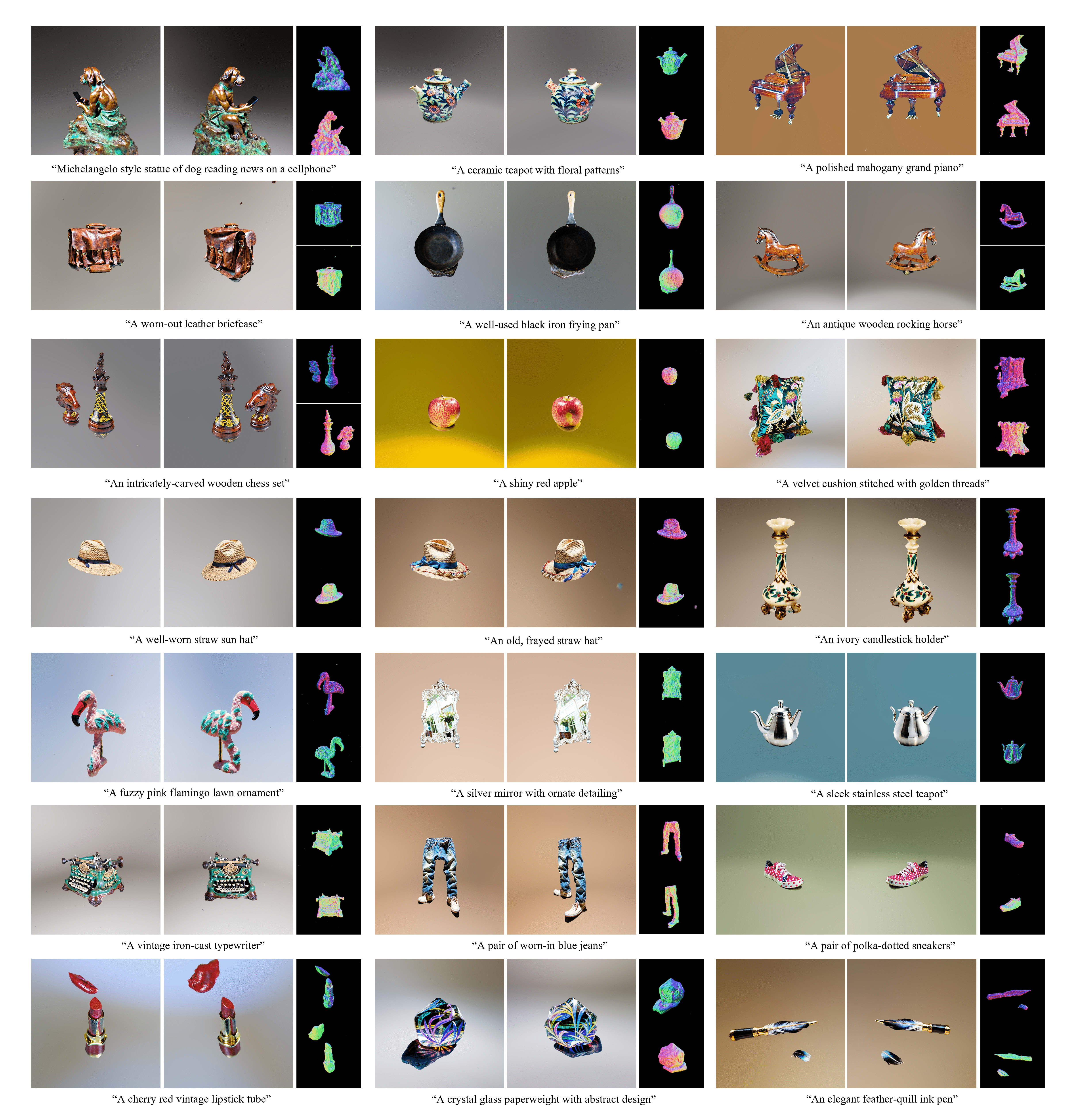} 
  \caption{Additional generations from our method.} 
  \label{fig:extra} 
\end{figure*}

Figure \ref{fig:extra} provide additional generations produced using our method. Our additional results are primarily intended to highlight the advantages of TBSD in terms of both texture details and shape fidelity. As shown, our method consistently generates high-quality outputs across prompts involving various materials and object categories, demonstrating its strong generalizability and high-performance ceiling. 

\subsection{Additional comparison with baselines}
\begin{figure*}[h!]
  \includegraphics[width=1\textwidth]{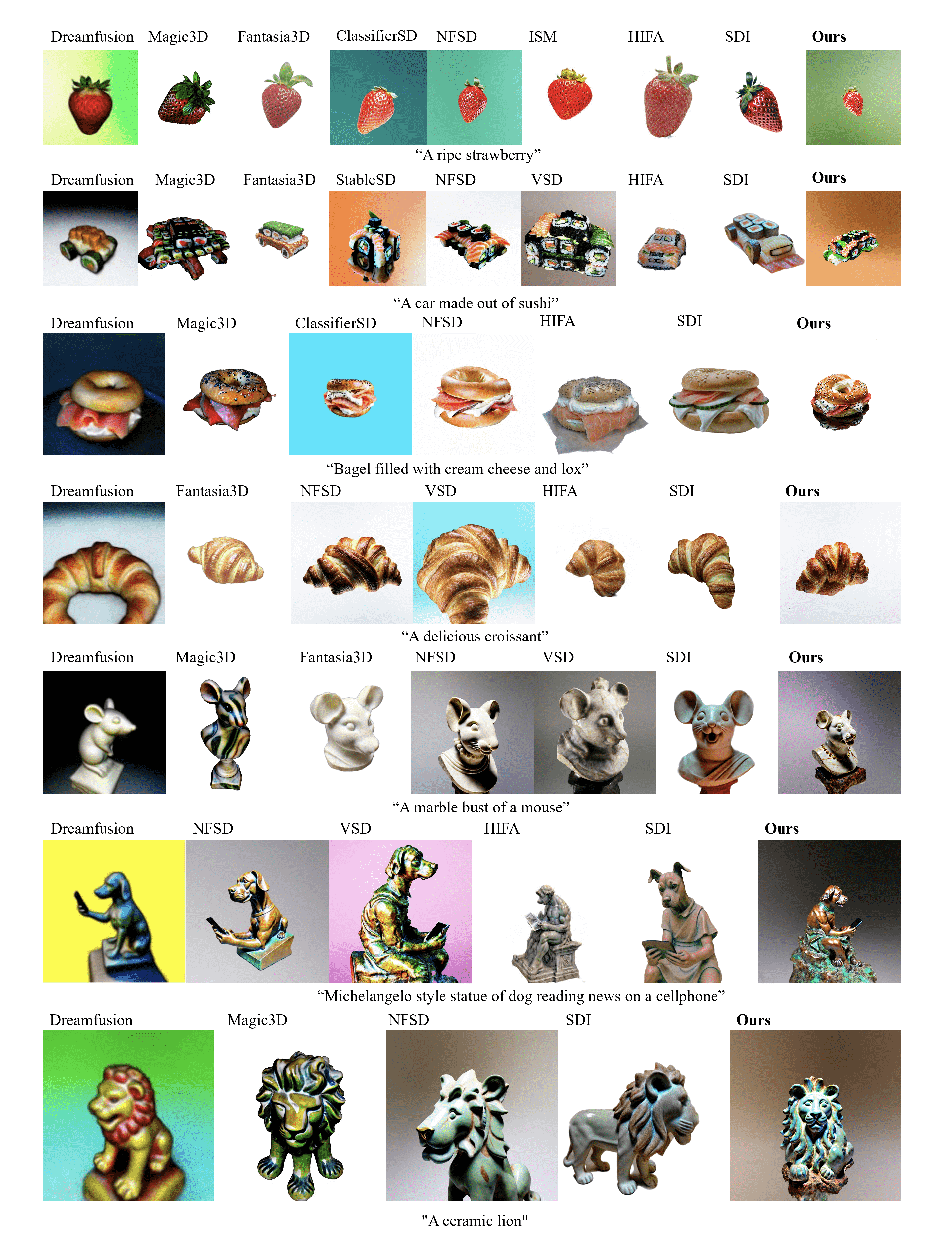} 
  \caption{Additional comparisons.} 
  \label{fig:compare1} 
\end{figure*}

\begin{figure*}[htbp]
  \includegraphics[width=1\textwidth]{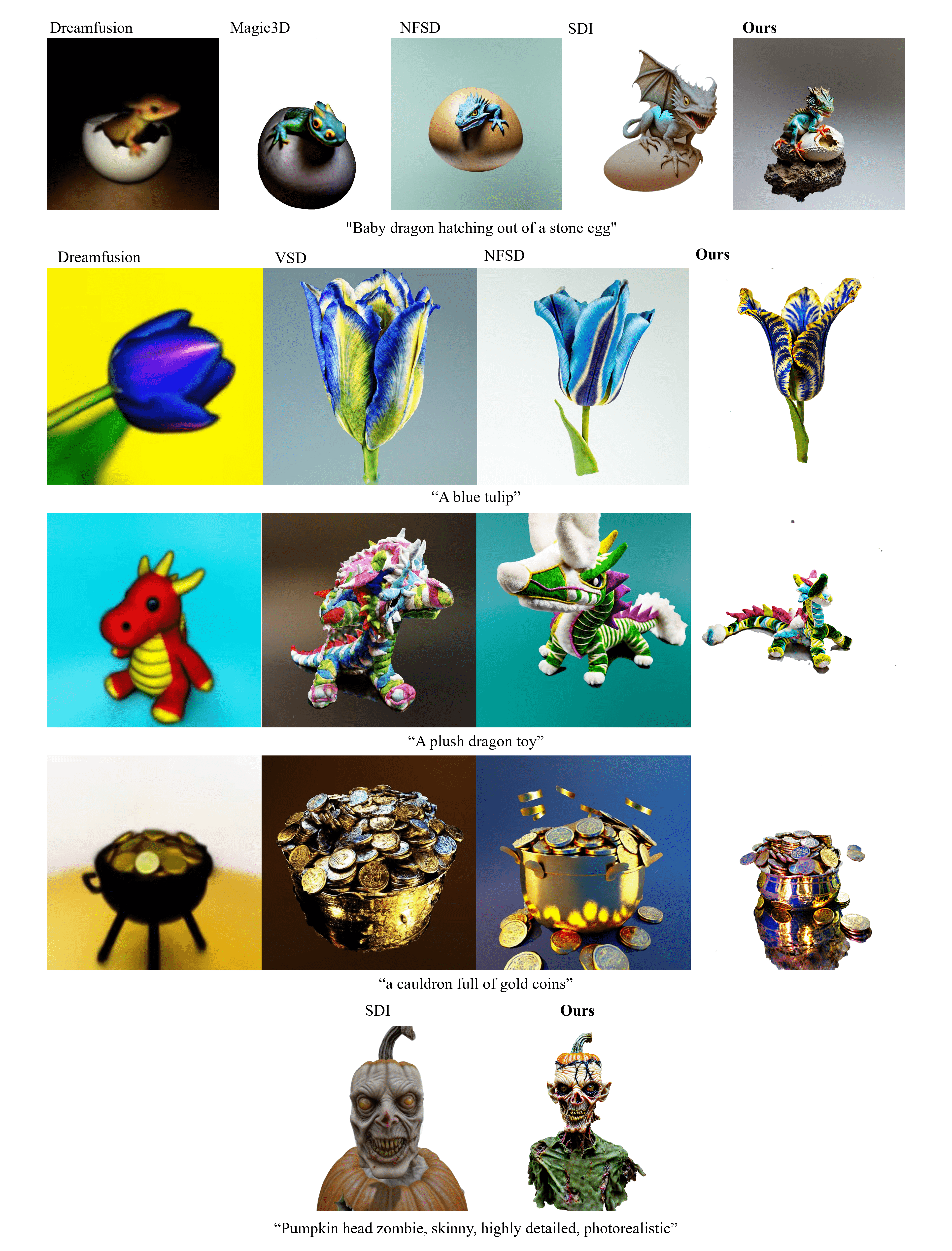} 
  \caption{Additional comparisons.} 
  \label{fig:compare2} 
\end{figure*}

We provide additional qualitative comparisons with other score distillation algorithms in Figures \ref{fig:compare1} to \ref{fig:compare2}. We compare with Dreamfusion, Magic3D, Fantasia3D, Stable Score Distillation (StableSD), Classifier Score Distillation (ClassifierSD), Noise-Free Score Distillation (NFSD), ProlificDreamer or Variational Score Distillation (VSD), Interval Score Matching (ISM), HiFA and SDI. It can be observed that our method achieves comparable or even superior results compared to other approaches. In particular, it clearly outperforms VSD (which relies on additional LoRA-based training) and SDI (which adopts a multi-step optimization strategy) in terms of color fidelity and texture details.

\section{Prompts used in the Qualitative evaluation}

{\fontsize{8pt}{13pt}\selectfont

\noindent ``A car made out of sushi''

\noindent``A delicious croissant''

\noindent``A small saguaro cactus planted in a clay pot''

\noindent``A plate piled high with chocolate chip cookies''

\noindent``A 3D model of an adorable cottage with a thatched roof''

\noindent``A marble bust of a mouse''

\noindent``A ripe strawberry''

\noindent``A rabbit, animated movie character, high detail 3d mode''

\noindent``A stack of pancakes covered in maple syrup''

\noindent``An ice cream sundae''

\noindent``A baby bunny sitting on top of a stack of pancakes''

\noindent``Baby dragon hatching out of a stone egg''

\noindent``An iguana holding a balloon''

\noindent``A blue tulip''

\noindent``A cauldron full of gold coins''

\noindent``Bagel filled with cream cheese and lox''

\noindent``A plush dragon toy''

\noindent``A ceramic lion''

\noindent``Tower Bridge made out of gingerbread and candy''

\noindent``A pomeranian dog''

\noindent``A DSLR photo of Cthulhu''

\noindent``Pumpkin head zombie, skinny, highly detailed, photorealistic''

\noindent``A shell''

\noindent``An astronaut is riding a horse''

\noindent``Robotic bee, high detail''

\noindent``A sea turtle''

\noindent``A tarantula, highly detailed''

\noindent``A DSLR photograph of a hamburger''

\noindent``A DSLR photo of a soccer ball''

\noindent``A DSLR photo of a white fluffy cat''

\noindent``A DSLR photo of a an old man''

\noindent``Renaissance-style oil painting of a queen''

\noindent``DSLR photograph of a baby racoon holding a hamburger, 80mm''

\noindent``Photograph of a black leather backpack''

\noindent``A DSLR photo of a freshly baked round loaf of sourdough bread''

\noindent``A DSLR photo of a decorated cupcake with sparkling sugar on top''

\noindent``A DSLR photo of a dew-covered peach sitting in soft morning light''

\noindent``A photograph of a policeman''

\noindent``A photograph of a ninja''

\noindent``A photograph of a knight''

\noindent``An astronaut''

\noindent``A photograph of a firefighter''

\noindent``A Viking panda with an axe''
}

\end{document}